\documentclass[letterpaper, 10 pt, conference]{ieeeconf}  % Comment this line out if you need a4paper

\IEEEoverridecommandlockouts                              % This command is only needed if 
\usepackage{amsthm}
\usepackage{xparse}
\usepackage{amsmath,bm}
\usepackage{amssymb,amsfonts}
\usepackage{times}
\usepackage{graphicx}
\usepackage{color}
\usepackage{caption}
\usepackage{float}
\usepackage{subfigure}
\usepackage{array}
\usepackage{multirow}
\usepackage{makecell}
\usepackage{xcolor}
\usepackage{mathtools}
\usepackage{booktabs} %调整表格线与上下内容的间隔
\usepackage{amsthm}
\usepackage{multirow}
\usepackage{mathrsfs}
\usepackage{array}
\usepackage{lineno,hyperref}
\usepackage{mathrsfs}
\usepackage{caption}
\usepackage{nomencl}
\usepackage{program}
\usepackage{hyperref}
\usepackage{algorithm}
\usepackage{algpseudocode}
\usepackage{setspace}
\usepackage{arydshln}
\usepackage{threeparttable}
\usepackage{notoccite}

\title{\LARGE \bf
DAMS-LIO: A Degeneration-Aware and Modular Sensor-Fusion LiDAR-inertial Odometry
}

\author{Fuzhang~Han$^{1}$\textsuperscript{*},
        Han~Zheng$^{2}$\textsuperscript{*},
		Wenjun~Huang$^{1}$,
		Rong~Xiong$^{1}$,
		Yue~Wang$^{1}$,
		and~Yanmei~Jiao$^{3}$ % stops a space
\thanks{*Both authors contributed equally to this work}% <-this % stops a space
\thanks{This work is supported by the National Key R$\&$D Program of China(2020YFB1313300), "Ling-Yan" Research and Development Project of Zhejiang Province of China(2023C03185) and Natural Science Foundation of Zhejiang Province(LGG21F030012). $^{1}$Fuzhang Han, Yue Wang, Wenjun Huang and Rong Xiong are with the State Key Laboratory of Industrial Control and Technology, Zhejiang University, Hangzhou, P.R. China. $^{2}$Han Zheng is with the College of Electrical Engineering, Zhejiang University, China. $^{3}$Yanmei Jiao is with the School of Information Science and Engineering, Hangzhou Normal University, Hangzhou 311121, China.}%
\thanks{Corresponding author, {\tt\small ymjiao@hznu.edu.cn}, Co-corresponding author, {\tt\small wangyue@iipc.zju.edu.cn.}}%
}

\begin{document}

\maketitle
\thispagestyle{empty}
\pagestyle{empty}

%%%%%%%%%%%%%%%%%%%%%%%%%%%%%%%%%%%%%%%%%%%%%%%%%%%%%%%%%%%%%%%%%%%%%%%%%%%%%%%%
\begin{abstract}
With robots being deployed in increasingly complex environments like underground mines and planetary surfaces, the multi-sensor fusion method has gained more and more attention which is a promising solution to state estimation in the such scene. The fusion scheme is a central component of these methods.
% The fusion scheme is crucial to the multi-sensor fusion method, which is the promising solution to state estimation in complex and extreme environments . 
In this paper, a light-weight iEKF-based LiDAR-inertial odometry system is presented, which utilizes a degeneration-aware and modular sensor-fusion pipeline that takes both LiDAR points and relative pose from another odometry as the
measurement in the update process only when degeneration is detected. Both the Cramer-Rao Lower Bound (CRLB) theory and simulation test are used to demonstrate the higher accuracy of our method compared to methods using a single observation. Furthermore, the proposed system is evaluated in perceptually challenging datasets against various state-of-the-art sensor-fusion methods. The results show that the proposed system achieves real-time and high estimation accuracy performance despite the challenging environment and poor observations.
\end{abstract}

%%%%%%%%%%%%%%%%%%%%%%%%%%%%%%%%%%%%%%%%%%%%%%%%%%%%%%%%%%%%%%%%%%%%%%%%%%%%%%%%
\section{Introduction}
Over the last decade, robotic systems have witnessed a significant increase in popularity and are used in increasingly complex applications like search and rescue \cite{erdelj2017help}, exploration, and mapping  \cite{pan2020gem}. However, accurate and robust state
estimation in these scenarios is not trivial due to the poorly
lit (e.g., fog, dust) and self-similar structures (e.g., tunnels,
long corridors), which leads to the lack of reliable visual and
geometry features. These problems pose unique challenges to
methods based on single sensor observations (e.g., incorrect
feature matching or degradation problems) \cite{jiao20192} \cite{jiao2021robust} \cite{shan2018lego}. Thus multi-modal sensor fusion has been widely deployed in such tasks \cite{zhao2021super} \cite{jing2022dxq}. Based on their design scheme, these approaches can be grouped into two main categories: tightly-coupled methods and loosely-coupled methods.

%As LiDAR can provide accurate long-range 3D measurements without relying on external light sources, it has been widely
%Accuracy localization in these scenarios usually  are not trival, lack of prominent perceptual   localization play a critial role in 
%rapidly 
%Over the last decade, LiDAR sensors have beening widely used in a broad spectrum of applications ranging from autonomous driving to industrial monitoring\cite{losch2018design}, for that can provide accurate long-range 3D measurements without relying on external light sources. However, LiDAR-based methods tend to suffer in  structure-less or self similar environments, like long straight tunnel and planar wall, giving rise to the problem of degeneration \cite{zhou2020LiDAR}. To handle that situation, some studies focus on  integration of the LiDAR sensor and the other sensors with complementary properties. Based on their design scheme, these approaches can be grouped into two main categories: tightly-coupled methods and loosely-coupled methods.
The former considers the error of different observations simultaneously, such as reprojection errors of visual features and point-to-plane error of Light Detection and Ranging (LiDAR) points \cite{shan2021lvi} \cite{zuo2019lic} \cite{lin2021r}. They show marked improvement in accuracy and robustness in most scenarios. But as for above mentioned extreme conditions, tightly-coupled methods may be susceptible to sensor failures since most of them only use a single estimation engine \cite{zhao2021super}. Moreover, although these methods have an analysis of the quality of different measurements (e.g., number of feature points), in some conditions, like dark but geometrically feature-rich places, their localization accuracy may be lower than those using single high-fidelity measurements because performing multi-sensor fusion all the time injects more noise to the estimator.

In comparison, loosely coupled methods are more robust in such perceptually-challenging conditions. Usually, they have a primary estimation engine (typically LiDAR-based methods) and regard the estimation result of different odometry as the initial value of scan-to-scan or scan-to-map \cite{palieri2020locus} \cite{khattak2020complementary} when they detect degeneration of the primary engine. These approaches provide a decent trade-off between accuracy and robustness and have witnessed great success in the recent DARPA Robotics Challenge \cite{rouvcek2019darpa}. However, the other odometry in these methods have no influence on the hessian matrix of the system, so they do not fully utilize the information of other odometry and the performance may be severely compromised in the case of a poor prior.
% Above mentioned methods perform multi-sensor fusion all the time, which may inject more noise into the estimator  and degrades the localization performance in case one of measurements is not well-conditioned. 
% Hence, some researchers have come up with algorithms based primarily on LiDAR odometry and integrating with other odometers when degeneration is detected, which provide a decent trade-off between accuracy and robustness and have witness great success in the recent DARPA Robotics Challenge\cite{rouvcek2019darpa}. 
\begin{figure}
	\setlength{\belowcaptionskip}{-0.6cm}   %调整图片标题与下文距离
	\centering \includegraphics[width=0.45\textwidth]{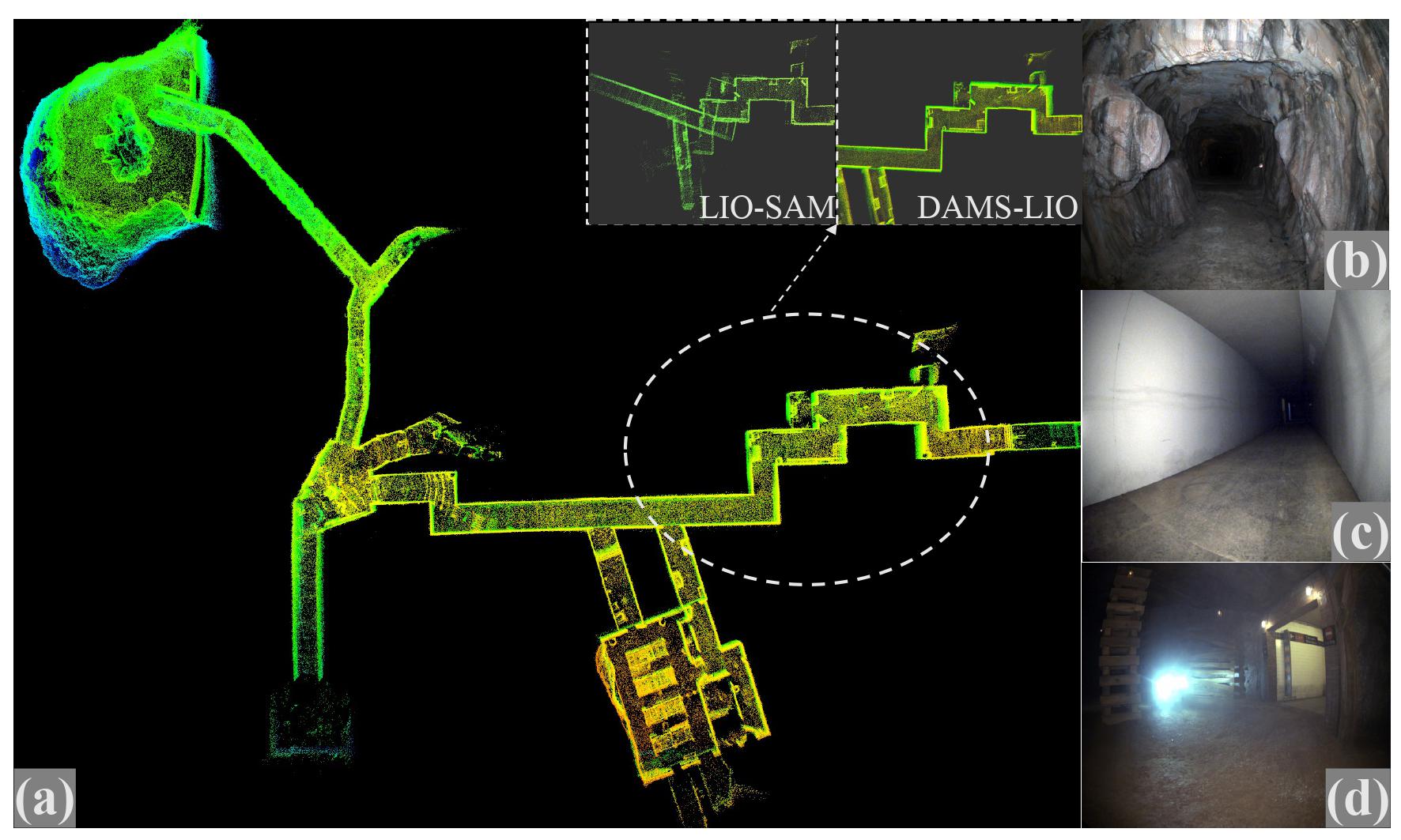}
	\captionsetup{font={footnotesize}}
	\caption{{(a) is the mapping result of DAMS-LIO in the CERBERUS DARPA Subterranean Challenge Datasets, which contains lots of challengeing environments as shown in (b)-(c) including darkness, long tunnel and textureless area. We compare our mapping results with the state-of-the-art LiDAR-inertial odometry, LIO-SAM, in the upper right corner of (a).}}\label{begin}
\end{figure}

% Motivated by the discussion above, we propose a degeneration-aware and modular sensor-fusion pipeline in the iterated extended Kalman filter framework (iEKF) \cite{xu2022fast}.
Motivated by the discussion above, we propose a degeneration-aware and modular sensor-fusion pipeline in the iterated extended Kalman filter framework (iEKF), drawing on the state-of-the-art LiDAR-inertial odometry FAST-LIO2 \cite{xu2022fast}. 
Following the loosely-coupled model, it works as LiDAR-inertial odometry in a well-condition and performs sensor-fusion with other odometry when degeneration is detected. The distinctive insight of our approach is that we take both LiDAR feature points and relative pose provided by other odometry as measurements to participate in the subsequent update in iEKF so that the other odometery information can be fully utilized while relying less on them.
% Craḿer-Rao
Theoretical analysis based on Cramer-Rao Lower Bound (CRLB) \cite{gorman1990lower} theorem demonstrates that integrating relative pose as measurements into the update process can achieve higher accuracy than as initial values in registration. 
% Simultaneously, we perform online calibration by taking extrinsic parameters of another odometry and LiDAR odometry into the state vector to facilitate the deployment of the system.

%that fully consider LiDAR and pose measurements provided by other odometry.

In summary, the contributions of this paper are listed as follows: 
\begin{itemize}
\item A lightweight degeneration-aware and modular sensor-fusion LiDAR-inertial odometry system (DAMS-LIO) is proposed, which performs robust and accurate state estimation in extreme environments and offers a marked advantage for the complex exploration tasks by robots with limited computing resources.
\item A novel sensor-fusion method to fully fuse the information of LiDAR and the other odometry is proposed, which takes both LiDAR points and relative pose from the other odometry as measurements in the update process only when degeneration is detected.
\item Theoretical analysis based on CRLB theorem is performed to quantify the performance and demonstrate the high accuracy of the proposed sensor-fusion method.
% \item We come up with a novel senor-fusion method to fully utilize the odometry information while circumvent introducing unnecessary loss of accuracy by degeneration-aware scheme that fuses the relative pose only when degeneration is detected.
% \item Theoretical analysis is performed to quantify the performance of our sensor-fusion algorithm.
% \item We proposed a lightweight modular sensor fusion pipline that fuse LiDAR and other odometry measurements, as well as online extrinsic calibration, that enables accurate real-time state estimation in extreme and perceptually-challenging environments and can be easily deployed on various robotic platforms.
\item Extensive experiments on simulation and real-world datasets validate the robustness and accuracy of our method.
\end{itemize}

\section{Related Works}
Since autonomous robots in extreme and unknown environments (e.g., underground or planetary exploration) is subject to many challenges and limitations, individual sensor modalities might fail (e.g., due to camera blackouts or degenerate geometries for LiDAR). Hence in recent years, there are several efforts have been made for the multi-sensor fusion method, which can be classified as either loosely coupled methods or tightly coupled methods.
\subsection{Tightly Coupled Method}
The tightly-coupled method typically incorporates the measurement of different sensors into the state optimization process. In the work of \cite{shan2021lvi}, LiDAR-inertial and visual-inertial systems are fused based on a tightly-coupled smooth and mapping framework, which can work independently when failure is detected in one of them or jointly in a well-condition. \cite{zuo2019lic} proposes an MSCKF-based LIC-fusion framework, which performs state estimation by IMU measurements, sparse visual and LiDAR features, and simultaneous online spatial and temporal calibration. Although these methods show marked improvement in the system's robustness and accuracy, they are susceptible to failure when the sensors are damaged and difficult to extend to other sensors.

\subsection{Loosely Coupled Method}
Zhang and Singh  \cite{zhang2018laser} propose V-LOAM, which utilizes the result of loosely-coupled Visual-Inertial odometry (VIO) prior to the initialization of the  LiDAR mapping system. \cite{palieri2020locus} comes up with a multi-sensor LiDAR-centric solution, LOCUS, which adds a health monitoring module to select a near-optimal prior to the LiDAR scan-matching optimization. These loosely-coupled approaches show more robust performance and higher resilience compared to tightly-coupled ones. However, their final performance still relies on laser scan alignment. Thereby these methods are still sensitive to the quality of LiDAR data. Hence, to fully utilize the information of the prior pose while retaining the advantages of the loosely-coupled method, we come up with a degeneration-aware and LiDAR-centric sensor fusion pipeline. It only receives pose measurements when LiDAR inertial odometry fails to make a trade-off between robustness and accuracy.
% Moreover, the online calibration is deployed so that the system can be convenient to be applied to various robot systems.
% \begin{figure}
% 	\setlength{\belowcaptionskip}{-0.6cm}   %调整图片标题与下文距离
% 	\setlength{\abovecaptionskip}{-0.2cm}
% 	\centering \includegraphics[width=0.38\textwidth]{fig/frame.pdf}
% 	\captionsetup{font={footnotesize}}
% 	\caption{{Illustration of each frames.}}\label{frame}
% \end{figure}
\section{System Overview}
The definition of each frame in our system is follows the typical LIO frame definitions.
% The definition of each frame in our system is illustrated in Fig. \ref{frame}. Following typical LIO frame definitions,
$\{L\}$ and $\{I\}$ are corresponding to the LiDAR and IMU frame respectively, while $\{G\}$ is a local-vertical reference frame whose origin coincides with the initial IMU position. $\{O\}$ represents the other odometry measurement, whose origin is denoted as $\{M\}$. $^L\bm{p}$ denotes the position of each point to the LiDAR frame.

The definition of some important and frequently used symbols are shown in Table \ref{symbol_definition}.
\begin{table}[h]
\centering
	\scriptsize
	\caption{
		Important Symbols Definition
	}
	\label{symbol_definition}
	%	\begin{adjustbox}{width=\columnwidth,center}
	% \resizebox{0.47\textwidth}{!}{}
		\begin{tabular}{cl} \hline \specialrule{0em}{1.5pt}{1.5pt}
			\textbf{Symbols} & \textbf{Definition}  \\ \midrule
			$t_k,t_{k+1}$ & The sample time of $\mathit{k}$-th and its next IMU measurement\\
			$\tau_i,\tau_{i+1}$ & The scan time of $\mathit{i}$-th and its next scan\\
			$I_k,I_i$ & The IMU frame at time $t_k$ and $\tau_{i}$ \\ 
			$L_k,L_{k+1}$ & The LiDAR frame at time $t_k$ and $t_{k+1}$ \\
			$O_k,O_{k+1}$ & The other odometry frame at time $t_k$ and $t_{k+1}$ \\
			$\mathbf{x},\hat{\mathbf{x}},\bar{\mathbf{x}}$ & The true, predicted and updated value of x \\
			$\mathbf{\hat{x}}^{\kappa}$& The $\kappa$-th update of $\bm{x}$ in the iterated Kalman filter\\
			$\tilde{\mathbf{x}}$ & The error between true x and its estimation value $\bar{\mathbf{x}}$\\
			$^{A}_{B}q,{^{A}p_{B}}$ &  \makecell{The rotation (represented by quaternion and its rotation \\matrix is $^{A}_{B}R$) and translation from frame A to B} \\
			\hline \specialrule{0em}{1.5pt}{1.5pt}
	\end{tabular}
	\vspace{-0.8cm}
	%	\end{adjustbox}
\end{table}
 \subsection{State Vector} The states estimated in our system containing current IMU states $\mathbf{x}_{I}$, the extrinsic parameters between LiDAR and IMU $^{I}\mathbf{x}_L$, and the transformation from other odometry frame to IMU frame $^{I}\mathbf{x}_O$. At time $t_k$, the state is written as:
\begin{flalign}
	\mathbf{x}_k =& \ [\bm{x}_{I,k}^{\top}\quad \bm{x}_L^{\top} \quad \bm{x}_{O}^{\top}]^{\top} & \\[1.5mm] \label{state vector}
	\mathbf{x}_{I_k} =&\ [^{G}_{I_k}\bar{\bm{q}}^{\top} \quad
	^G\bm{p}^{\top}_{I_k} \quad
	^G\bm{v}^{\top}_{I_k}  \quad	
	\bm{b}^{\top}_{g_k} \quad
	\bm{b}^{\top}_{a_k} \quad
	\bm{g}_{k}\ ]^{\top}
	& \\[1.5mm]
	\mathbf{x}_{L} =&\ [^{I}_{L}\bar{\bm{q}}^{\top}  \quad ^{I}\bm{p}^{\top}_{L}]^{\top} , \quad	\mathbf{x}_{O} =\ [^{I}_{O}\bar{\bm{q}}^{\top}  \quad ^{I}\bm{p}^{\top}_{O}]^{\top} 
\end{flalign}
$^G\bm{v}_{I}$ is the velocity of IMU in global frame, $\bm{b}_{g,k}$ and $\bm{b}_{a,k}$ represent the gyroscope and accelerometer biases respectively, $\bm{g}$ is the gravity vector in frame $\{G\}$, $\mathbf{x}_L = \{^{I}_{L}\bar{\bm{q}},\ ^{I}\bm{p}_{L}\}$ and $\mathbf{x}_O = \{^{I}_{O}\bar{\bm{q}},\ ^{I}\bm{p}_{O}\}$ denotes the transformation from other odometry frame $\{O\}$ and LiDAR frame $\{L\}$ to IMU frame $\{I\}$. 

\subsection{IMU Propagation}
Since measurements of IMU are affected by bias $\bm{b}$ and zero-mean Gaussian noise $\bm{n}$ \cite{yang2019degenerate}, they can be modeled as:
\begin{eqnarray}\label{IMU measurement model}
	\bm{\omega}_{m}(t) &=& ^I\bm{\omega}(t) + \bm{b}_g(t) + \bm{n}_g(t) \\
	\bm{a}_m(t) &=& {^{I(t)}_G\bm{R}(^G\bm{a}_I(t) + {^G\bm{g}})} + \bm{b}_a(t)+ \bm{n}_a(t)
\end{eqnarray}
where $\bm{\omega}_m(t)$ and $\bm{a}_m(t)$ are the raw measurement data. $^I\bm{\omega}(t)$ is the angular velocity of IMU in local frame $\{I\}$. $^G\bm{g}$ and $^G\bm{a}_I(t)$ are the acceleration of gravity and IMU expressed in the global frame.
% The IMU kinematics \cite{trawny2005indirect} can be described as:
% \begin{equation}
% 	\begin{aligned}\label{IMU kinematics}
% 		^I_G\dot{\bar{\bm{q}}}(t) &= \frac{1}{2}\bm{\Omega}(^I\bm{\omega}(t))^I_G{\bar{\bm{q}}}(t) \\
% 		^G\dot{\bm{p}_{I}}(t) &= {^G\bm{v}_I(t)}, \quad ^G\dot{\bm{v}}_I(t) = {^G\bm{a}(t)} \\
% 		\dot{\bm{b}_g}(t) &= \bm{n}_{wg}(t), \quad \dot{\bm{b}_a}(t) = \bm{n}_{wa}(t)
% 	\end{aligned}
% \end{equation}
% where $\bm{\Omega}(\bm{\omega}) = \begin{bmatrix}- \lfloor {\bm{\omega}} \times \rfloor & \bm{\omega} \\ -\bm{\omega}^{\top} & 0\end{bmatrix}$ and $\lfloor {\bm{\omega}} \times \rfloor$ is the skew-symmetric matrix. Besides, the value of $\mathbf{x}_L$ and $\mathbf{x}_O$ are independ of time. Hence, we can obtain the propagated value $\hat{\bm{x}}_{k+1}$ based on （\ref{IMU kinematics}) and $\hat(\bm{x})$.
The IMU kinematics is the same as \cite{li2013high}, to keep our presentation concise, we do not repeat the description here.
To propagate the covariance matrix from time $t_k$ to $t_{k+1}$, we have the generation format of the linearized discrete-time model, following \cite{mourikis2007multi} as:
\begin{align}
	\label{discrete-time-model}  \tilde{\bm{x}}_{k+1} &= \bm{\Phi}_{k}\tilde{\bm{x}}_k + \bm{G}\bm{n}_k 
\end{align}
where $\bm{\Phi}_{k}$ is the linearized system state transition matrix, $\bm{n}_k = [\bm{n}_{g} \quad\bm{n}_{wg}\quad\bm{n}_{a}\quad\bm{n}_{wa}]$ is the system noise. The error state is defined as $\tilde{\bm{x}} = \bm{x} - \hat{\bm{x}}$ for all variables except for quaternion, which is defined by the relation $q = \hat{q} \otimes \delta q$. Same as \cite{li2013high}, the symbol $\otimes$ means quaternion multiplication, and the error quaternion is defined as $\delta q \simeq [\frac{1}{2}\bm{\delta \theta}^{\top}\quad 1]^{\top}$. 

Denoting the covariance of $\bm{n}_k$ as $\bm{Q}_k$, then we can propagate the state covariance from $t_k$ to $t_{k+1}$ as:
\begin{align} 
    \bm{P}_{k+1} = \bm{\Phi}\bm{P}_{k}\bm{\Phi}^{\top} +\bm{G}\bm{Q}_k\bm{G}^{\top}
\end{align}
where $\bm{\Phi} = {\rm diag}(\bm{\Phi}_{I}, \bm{\Phi}_{O})$ and $\bm{G} = [\bm{G}^{\top}_{I},\bm{G}^{\top}_{O}]^{\top}$. $\bm{\Phi}_{I}$ and $\bm{G}_{I}$ indicate the part related to the variable other than the extrinsic between other odometry and IMU, which are the same as the definition in \cite{xu2022fast}. Moreover $\bm{\Phi}_{O} = \bm{I}_{6\times6}$ and $\bm{G}_{O} = \bm{0}_{6\times12}$.
\begin{figure}
	\setlength{\belowcaptionskip}{-0.6cm}   %调整图片标题与下文距离
	\setlength{\abovecaptionskip}{-0.3cm}
	\centering \includegraphics[width=0.43\textwidth]{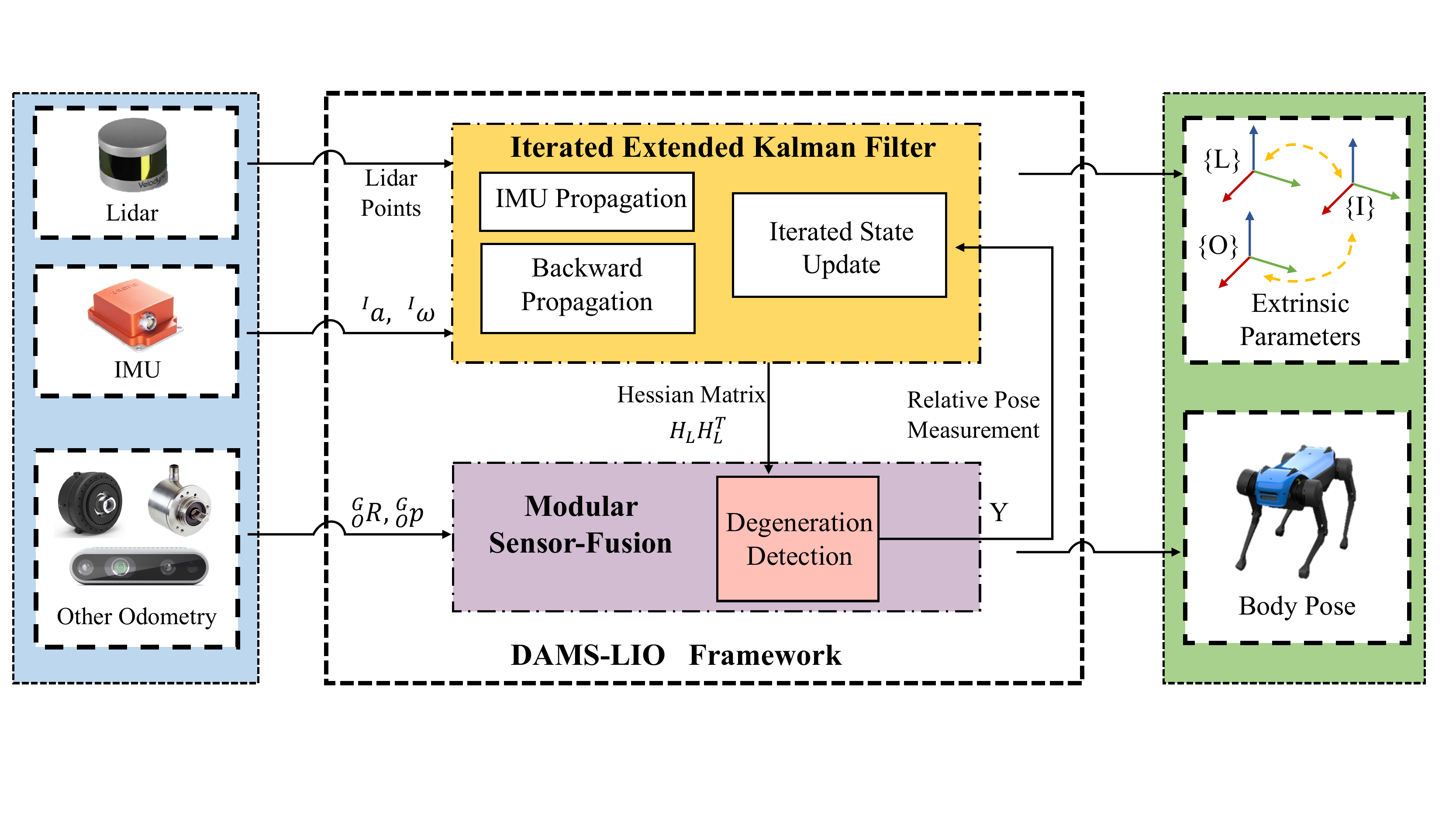}
	\captionsetup{font={footnotesize}}
	\caption{{Framework of the proposed DAMS-LIO.}}\label{framework}
\end{figure}

\subsection{Measurement Model}
1) \textbf{LiDAR measurement}:
As for the measurements from LiDAR, we model them as in \cite{xu2022fast}. After motion compensation for the scan sampled at time $\tau_i$, we define the $\mathit{j}$-th point at the local LiDAR frame as $^{L}\bm{p}_j$. Through backward propagation, the point can be changed to a scan-end measurement corresponding to the IMU measurement at $t_k$. Meanwhile, since each point should lie on a small plane patch in the map, we can get
\begin{small}
\begin{align} \label{LiDAR model}
    \bm{0} = {^{G}}\bm{u}_j^{\top}(^{G}_{I_k}\bm{R} ({^{I}_L\bm{R}}(^{L}\bm{p}_j + {^{L}\bm{n}_j})+{^{I}\bm{p}_L})+{^{G}\bm{p}_{I_k}} -{^{G}\bm{q}_j} )
\end{align}
\end{small}
where $^{G}\bm{q}_j$ is the point on the small plane and $^{G}\bm{u}_j$ is the normal of the plane. $^{I}\bm{R}_L$ and $^{G}\bm{R}_{I_k}$ are the rotation matrix corresponding to the $^{I}_{L}\bm{q}$ and $^{G}_{I_K}\bm{q}$. $^{L}\bm{n}_j$ is the ranging and beam-directing noise of the point $^{L}\bm{p}_j$. (\ref{LiDAR model}) can also be summarized in a more compact from as
\begin{equation}\label{LiDAR h}
    {\bm{z}_l} = \bm{0} = {\bm{h}_l}(\bm{x}_k,{^{L}\bm{p}_j + {^{L}\bm{n}_j}}) 
\end{equation}
To linearize the measurement model for the update, we approximate the measurement model by its first order approximation at $\hat{\bm{x}_k}$ as
\begin{equation} \label{LiDAR r}
        \bm{r}_l = \bm{0} - {\bm{h}_l}(\bm{x}_k,{^{L}\bm{p}_j}) \simeq H_L {\bm{\tilde{x}_k}} + {\bm{v}_l}
\end{equation}
where ${\bm{v}_l} \in \mathit{N}(\bm{0},{\bm{R}_l})$ is the noise of Gaussian distribution due to the raw measurement noise ${^{L}\bm{n}_j}$. $H_L$ is the jacobian matrix of residual $\bm{r}_l$  for error state as ${\bm{\tilde{x}_k}}$, which is given as
\begin{small}
\begin{align}
    H_L &= {^{G}\bm{u}_j^{\top}}\begin{bmatrix}H_{L1}
  \ \bm{I}_3\ \bm{0}_{3\times12}& H_{L2}\quad ^G\hat{\bm{R}}_{I_k}\quad\bm{0}_{3\times6}
\end{bmatrix} \label{LiDAR jacobian1}\\
    H_{L1} &= {^G\hat{\bm{R}}_{I_k}}\lfloor {^L\bm{p}_j+{^I\hat{\bm{p}}_L}}\times\rfloor, H_{L2} = -{^G\hat{\bm{R}}_{I_k}}{^I\hat{\bm{R}}_{L}}\lfloor{^L\bm{p}_j}\times\rfloor \label{LiDAR jacobian2}
\end{align}
\end{small}
2) \textbf{The other odometry measurement}:
 Due to the different publish frequency of each odometry, we perform the linear interpolation to obtain the other odometry poses with the estimated state at time $t_k$ and $t_{k+1}$ and denote their frame as $O_{k-1}$ and $O_k$. We define the transformation matrix, $^{A}T_B$, from A to B as $^{A}T_B = [ ^{A}\bm{R}_{B} , ^{A}\bm{p}_{B} ;
    \bm{0}_{3\times3} , 1 ]$. Then, we have the transformation relationship as 
%  \begin{align} \label{transformation definition}
%     ^{A}T_B = \begin{bmatrix} ^{A}\bm{R}_{B} & ^{A}\bm{p}_{B} \\
%     \bm{0}_{3\times3} & 1
%     \end{bmatrix}
%  \end{align}
 \begin{align}\label{transformation definition}
     ^{O_{k-1}}T_{O_{k}} = (^{G}T_{I_{k-1}} {^{I}T_{O}})^{-1}(^{G}T_{I_{k}} {^{I}T_{O}})
 \end{align}
 where $^{O_{k-1}}T_{O_{k}} = {^{G}T_{O_{k-1}}^{-1}}{^{G}T_{O_{k}}}$ is the relative pose measurement calculated by other odometry. Hence based on (\ref{transformation definition}), we can easily get the measurement model $\bm{z}_O= \begin{bmatrix} \bm{z}_r \\ \bm{z}_p\\\end{bmatrix}$ as
 \begin{small}
%  \begin{align}
%       \bm{z}_r &= {\bm{h}_r}({\bm{x}_k},{\bm{x}_{k-1}}) = ({^{G}\bm{R}_{I_{k-1}}}{^{I}\bm{R}_O})^{\top}{{^{G}\bm{R}_{I_{k}}}{^{I}\bm{R}_O}} \label{odometry_meas1}\\
%      \begin{split}
%      \label{odomety_meas2}
%       \bm{z}_p &= {\bm{h}_p}({\bm{x}_k},{\bm{x}_{k-1}}) = ({^{G}\bm{R}_{I_{k-1}}}{^{I}\bm{R}_O})^{\top}(({^{G}\bm{R}_{I_{k}}} - {^{G}\bm{R}_{I_{k-1}}}){^{I}\bm{p}_O} \\  & +{^{G}\bm{p}_{I_{k}}} - {^{G}\bm{p}_{I_{k-1}}})
%      \end{split}
%  \end{align}
 \begin{align}
      \bm{z}_r &= {^{O_{k-1}}\bm{R}{_G}}{{^{G}\bm{R}_{I_{k}}}{^{I}\bm{R}_O}} \label{odometry_meas1}\\
     \begin{split}
     \label{odomety_meas2}
      \bm{z}_p &= {^{O_{k-1}}\bm{R}{_G}}(({^{G}\bm{R}_{I_{k}}}{^{I}\bm{p}_O} - {^{G}\bm{R}_{I_{k-1}}}{^{I}\bm{p}_O}) +{^{G}\bm{p}_{I_{k}}} - {^{G}\bm{p}_{I_{k-1}}})
     \end{split}
 \end{align}
\end{small}
%  {^{O_{k-1}}\bm{R}_{O_{k}}}
%  {^{O_{k-1}}\bm{p}_{O_{k}}}
where ${^{O_{k-1}}\bm{R}{_G}} = ({^{G}\bm{R}_{I_{k-1}}}{^{I}\bm{R}_O})^{\top}$ . We assume the state at $t_{k-1}$ is known, thus similar to (\ref{LiDAR h}) and (\ref{LiDAR r}), we have the residual of odometry measurement as $\bm{r}_O = \begin{bmatrix}\bm{r}_r\\ \bm{r}_p\end{bmatrix} \simeq \begin{bmatrix} H_{O_r} {\bm{\tilde{x}_k}} + {\bm{v}_r}\\H_{O_p} {\bm{\tilde{x}_k}} + {\bm{v}_p} \end{bmatrix}$, where $\bm{r}_r$\footnote{For rotation the minus operation is defined in Lie group} and $\bm{r}_p$ are the rotation and translation errors calculated like (\ref{LiDAR r}), ${\bm{v}_r} \in \mathit{N}(\bm{0},{\bm{R}_r})$ and ${\bm{v}_p} \in \mathit{N}(\bm{0},{\bm{R}_p})$ are the corresponding Gaussian noise.
% \begin{equation}
%     \bm{r}_O = \begin{bmatrix}\bar{\bm{z}_r} - {\bm{h}_r}({\bm{\hat{x}}_k},{\bm{x}_{k-1}}) \\ \bar{\bm{z}_p} - {\bm{h}_p}({\bm{\hat{x}}_k},{\bm{x}_{k-1}})\end{bmatrix} \simeq \begin{bmatrix} H_{O_r} {\bm{\tilde{x}_k}} + {\bm{v}_r}\\H_{O_p} {\bm{\tilde{x}_k}} + {\bm{v}_p} \end{bmatrix}
% \end{equation}
% \begin{equation}
%     \bm{r}_O \simeq \begin{bmatrix} H_{O_r} {\bm{\tilde{x}_k}} + {\bm{v}_r}\\H_{O_p} {\bm{\tilde{x}_k}} + {\bm{v}_p} \end{bmatrix}
% \end{equation}
Thus, based on (\ref{odometry_meas1}) and (\ref{odomety_meas2}), we have the jacobian matrix of rotation and translation residual for error state as
\begin{align}
    H_{O_r} &=\begin{bmatrix} ^I\hat{\bm{R}}_O^{\top} & \bm{0}_{3\times3}& \bm{0}_{3\times18}& H_{O_{r1}} & \bm{0}_{3\times3}
\end{bmatrix} \\
    H_{O_p} &=\begin{bmatrix} H_{O_{p1}} &    
  H_{O_{p2}}
&  \bm{0}_{3\times18}& 
H_{O_{p3}}
& H_{O_{p4}}
\end{bmatrix}
\end{align}
where
\begin{align}
    H_{O_{r1}} &=\bm{I}_3 - {{^I}\hat{\bm{R}}{^{\top}}_O}  {^G}\hat{\bm{R}}{^{\top}}_{I_k} {^G}\hat{\bm{R}}_{I_{k-1}}  {^I}\hat{\bm{R}}_O  \\
    H_{O_{p1}} &=-^I\hat{\bm{R}}{^{\top}}{_O} {^G}\hat{\bm{R}}{^{\top}}{_{I_{k-1}}}  {^G}\hat{\bm{R}}_{I_k}\lfloor{^I}\hat{\bm{p}}_O\times\rfloor \\
    H_{O_{p2}} &= ({^G}\hat{\bm{R}}_{I_{k-1}}{^I\hat{\bm{R}}{_O}})^{\top}\\
    H_{O_{p3}} &=\lfloor {^I}\hat{\bm{R}}^{\top}_O  {^G}\hat{\bm{R}}{^\top}{_{I_{k-1}}} \hat{\bm{P}}_1 \times\rfloor
\end{align}
\begin{align}
    \hat{\bm{P}}_1 &=({^G}\hat{\bm{R}}_{I_k}-{^G}\hat{\bm{R}}_{I_{k-1}}) {^I\hat{\bm{p}}_O} + {^G\hat{\bm{p}}_{I_k}} -{^G\hat{\bm{p}}_{I_{k-1}}}\\
    H_{O_{p4}} &=({^G}\hat{\bm{R}}{_{I_{k-1}}}{^I\hat{\bm{R}}{_O}})^{\top} ({^G}\hat{\bm{R}}{_{I_k}} - {^G}\hat{\bm{R}}_{I_{k-1}} )
\end{align}
\subsection{Degeneration-Aware Update}
Following \cite{xu2022fast}, we have the following error state:
\begin{equation} \label{prior}
    \bm{x}_k \boxminus \hat{\bm{x}}{_k} = ({\hat{\bm{x}}^{\kappa}}_k \boxplus {\tilde{\bm{x}}^{\kappa}}_k )\boxminus {\hat{\bm{x}}_k} = {\hat{\bm{x}}^{\kappa}}\boxminus {\hat{\bm{x}}_k} + {\bm{M}^{\kappa}} {\tilde{\bm{x}}^{\kappa}}_k
\end{equation}
where $\boxplus/ \boxminus$ means plus and minus operators in Lie group \cite{sola2018micro}. ${\bm{M}^{\kappa}}$ is partial differentiation of $({\hat{\bm{x}}^{\kappa}}_k \boxplus {\tilde{\bm{x}}^{\kappa}}_k )\boxminus {\hat{\bm{x}}_k}$ with respect to ${\tilde{\bm{x}}^{\kappa}}_k$ evaluated at zeros:
\begin{equation}
    \begin{split}
    {\bm{M}^{\kappa}} = {\rm diag}(\bm{A}({\delta{^G{\bm{\theta}}_{I_k}}})^{-\top},\bm{I}_{15\times15},\bm{A}({\delta{^I{\bm{\theta}}_{L_k}}})^{-{\top}}, \\
    \bm{I}_{3\times3},\bm{A}({\delta{^I{\bm{\theta}}_{O_k}}})^{-{\top}},\bm{I}_{3\times3})
    \end{split}
\end{equation}
where $\bm{A}(\cdot)$ is defined in \cite{he2021kalman} and ${\delta{^X{\bm{\theta}}_{Y}}} = {^X{\hat{\bm{R}}^{\kappa}}_Y}\boxminus{^X{\bm{\hat{R}}}_Y}$.
Based on (\ref{prior}) and taking it into the first-order approximation measurement model mentioned above, the problem can be summarized as:
\begin{equation}
    \underset{{\tilde{\bm{x}}^{\kappa}}_k}{{\rm min}}(\parallel {\hat{\bm{x}}^{\kappa}}\boxminus {\hat{\bm{x}}_k}\parallel^2_{\hat{\bm{P}}_k^{-1}} + \sum\parallel \bm{z}^{\kappa} + {H^{\kappa}}{{\tilde{\bm{x}}^{\kappa}}_k}  \parallel^2_{\bm{R}^{-1}})
\end{equation}

We exploit the agile update scheme to avoid the noise of pose measurement degrading the LiDAR odometry in well-condition and robustifying the odometry in degeneration conditions.
% as shown in Algorithm \ref{update algorithm}.
Theoretically, the eigenvalues corresponding to the degeneration dimensions are precisely zero. However, in practice,  they are typically a small value in the actual calculation due to the noise of data and limited computational accuracy. Hence, we first calculate the eigenvalues $\{\lambda_i\}$ of the Hessian matrix $\bm{H}^{\top}_L\bm{H}_L$ and refer to the heuristic method in \cite{ding2021degeneration} to determine how well the geometry feature of the scene. If the eigenvalue is smaller than the thread, we can infer the existence of degeneration. Usually, the setting of this value depends on the user's experience, so it may need to be adjusted when encountering different scenarios.

If LIO is in well condition and assume we have m LiDAR measurements, then ${{\bm{z}^\kappa}} = \bm{z}^{\kappa}_l=[{{\bm{z}^{\kappa}_{l1}}},\dots,{{\bm{z}^{\kappa}_{lm}}}]^{\top}$, $H ={H^{\kappa}_L}=[H^{\kappa}_{l1},\dots, H^{\kappa}_{lm}]^{\top}$ and $\bm{R}=\bm{R}_L={\rm diag}(\bm{R}_{l1},\dots,\bm{R}_{lm})$. Otherwise, if degeneration is detected, ${{\bm{z}^\kappa}} = [\bm{z}^{\kappa}_l,{\bm{z}^{\kappa}_r} ,{{\bm{z}^{\kappa}_p}}]^\top$, $H = [{H^{\kappa}_L} ,{H^{\kappa}_{O_r}} ,{H^{\kappa}_{O_p}} ]^\top$ and $\bm{R} ={\rm diag}(\bm{R}_L,\bm{R_{O_r}},\bm{R_{O_p}})$.
Then we can perform iterated Kalman filter update the same in \cite{xu2022fast}.
\section{Cramér–Rao Lower Bound Theorem}
In this section, we furnish the reader with further insight into the proposed fusion method and compare the accuracy with other purely LiDAR-based methods using the Cramér-Rao Lower Bound (CRLB). The CRLB is a lower bound on the variance of an estimator, which is often used to evaluate the performance of the data fusion method \cite{domhof2017multi} \cite{blanc2007data}. Following \cite{kowalski2019crlb}, the CRLB is calculated by taking the inverse of the Fisher information matrix as:
\begin{equation} \label{crlb_formation}
    CRLB = {\mathit{\bm{J}}^{-1}} = {{\bm{H}^{\top}}{\bm{R}^{-1}}{\bm{H}}^{-1}}
\end{equation}
where $\bm{H}$ and $\bm{R}$ are corresponding to the Jacobian and covariance matrix of measurement in section \uppercase\expandafter{\romannumeral3}. The smaller CRLB, the higher accuracy the system can achieve in theory.
For the needs of analysis, the following lemmas \cite{horn2012matrix} are needed.

\textit{Lemma 1}: The inverse of a positive definite matrix is also positive definite

\textit{Lemma 2}: A real symmetric matrix A is positive definite if there exists a real nonsingular matrix B such that $A=BB^{\top}$

\textit{Lemma 3}: Let A be a positive definite matrix and B be a $m\times n$ real matrix. ${B^{\top}AB}$ is positive definite if the rank for B:$\mathit{r}(B)=n$

Based on the theory mentioned above, we then calculate and compare the CRLB of different methods separately.

1) \textbf{Purely LiDAR-based method}:
Since the purely LiDAR-based method does not use odometry pose measurements, the extrinsic parameters related to odometry are removed from the state vector. Moreover, for simplicity of calculation, we ignore the variable whose corresponding part in $H$ is a zero vector. Thus based on (\ref{LiDAR jacobian1}) and (\ref{LiDAR jacobian2}), the measurement Jacobian matrix of purely LiDAR-based method is rewritten as
% \begin{align} \label{purely-LiDARH}
%     \bm{H}_{li} &= {^{G}\bm{u}_j^{\top}}\begin{bmatrix}\bm{H}_{L1}
%   & \bm{I}_3 & \bm{H}_{L2}\quad ^G_{\bm{I}_k}\hat{\bm{R}} \end{bmatrix} \\
%   &=\begin{bmatrix}\bm{H}_{pose}&\bm{H}_{extrinsic}\end{bmatrix}　 \nonumber
% \end{align}

\begin{small}
\begin{equation} \label{purely-LiDARH}
    \bm{H}_{li} = {^{G}\bm{u}_j^{\top}}\begin{bmatrix}\bm{H}_{L1}
  \; \bm{I}_3 \; \bm{H}_{L2} \; ^G_{\bm{I}_k}\hat{\bm{R}} \end{bmatrix}  =\begin{bmatrix}\bm{H}_{pose}\quad\bm{H}_{extrinsic}\end{bmatrix}
\end{equation}
\end{small}

We substitute (\ref{purely-LiDARH}) into (\ref{crlb_formation}) and represent it in the form of a chunking matrix as follows:
\begin{equation}\label{purely-LiDARJ}
    \mathit{\mathbf{J}}_{li} = \begin{bmatrix}
        \bm{U} & \bm{B} \\ \bm{B}^{\top} & \bm{C}\\
    \end{bmatrix}
\end{equation}
Since covariance matrix of LiDAR noise is a real symmetric matrix and the rank of $H_{li}$ is equal to its row, we can reason that $\mathit{\mathbf{J}}{_{li}}^{-1}$ is positive define matrix, which is denoted as $\mathit{\mathbf{J}}{_{li}}^{-1} > 0$. Then, based on the inverse formula for chunking matrix, we can easily obtain the part corresponding to the estimated pose as 
\begin{small}
\begin{equation} \label{crlb1}
    {\rm CRLB}_{li} = ({\bm{U} - \bm{B}{\bm{C}^{-1}}\bm{B}^{\top}})^{-1}
\end{equation}
\end{small}

% \begin{table*}\small
\begin{table*}\small
\caption{Accuracy Comparison Results of Each Method Operating on Various Challenge Environments}\label{ate compare}
\centering
\begin{tabular}{lcccccccc:cccccc} 
\hline\hline
Dataset & \multicolumn{8}{c}{CERBERUS} & \multicolumn{6}{c}{M2DGR}                                    \\ 
% \hline
          & \multicolumn{2}{c}{anymal1} & \multicolumn{2}{c}{anymal2} & \multicolumn{2}{c}{anymal3} & \multicolumn{2}{c}{anymal4} & \multicolumn{2}{c}{gate01} & \multicolumn{2}{c}{gate03} & \multicolumn{2}{c}{street03}  \\ 
\hline
  & max & mean  & max & mean                  & max & mean                  & max & mean                   & max & mean            & max & mean            & max & mean             \\ 
\hline
LOCUS     & 1.61 & 0.49 & -\tnote{１}  &      -         &      1.22    &  0.26 &  2.69 & 0.67 &  13.09      & 6.28 &  20.31   & 6.56  &  -    &  -  \\
LVI-SAM   & 8.18 &      1.11     & 1.86    & 0.44       &  2.87 &    0.45&   1.20&  0.36      & 10.01 &     4.28    & 22.93 &  9.06  & 19.14    &   9.67 \\
LIO-SAM   &  23.9 &   10.3  &  -   & -   & -  &     -  &    23.1  &     8.61   &13.27  &    6.68 & 12.27  &     3.30    & 12.86  &  7.52    \\
VINS-MONO &  5.93   &  2.34    &    11.20  &    1.55   &4.40   &    1.48   &2.00 & 0.72              & 6.92 & 3.75  & 12.75  & 8.21   &14.46 &  4.61    \\
Fast-LIO2 & \textbf{0.48} & \textbf{0.14}  & 0.41  & \textbf{0.14}   & -  &  -      &  22.75 & 0.55    & 4.55 & 2.29    & 8.37  & 2.93            & 5.22 & 0.72               \\
DAMS-LIO  & 0.67  & 0.20  & \textbf{0.35} & \textbf{0.14}  &  \textbf{0.73} & \textbf{0.17}  &  \textbf{1.09} &  \textbf{0.27} &  \textbf{2.73}    & \textbf{0.94}  & \textbf{3.38}   &  \textbf{2.24} & \textbf{1.69} &  \textbf{0.47}        \\
\hline\hline
\end{tabular}
 \begin{tablenotes}
    \footnotesize
    \item[1] "-" means that the method fails. The units for all values are meters.
\end{tablenotes}
\vspace{-0.5cm} 
\end{table*}

2) \textbf{Pose-fusion method}:
Following the process mentioned above, we have the measurement Jacobian matrix as 
% \begin{equation}
%     \bm{H}_{pf} = \begin{bmatrix}{^{G}\bm{u}_j^{\top}}\bm{H}_{L1}
%   & {^{G}\bm{u}_j^{\top}} & {^{G}\bm{u}_j^{\top}}\bm{H}_{L2}＆ {{^{G}\bm{u}_j^{\top}}}^G_{\bm{I}_k}\hat{\bm{R}}& \bm{0}_{3\times3}& \bm{0}_{3\times3} \\ {^I\hat{\bm{R}}_O^{\top}} & \bm{0}_{3\times3}& \bm{0}_{3\times3}& \bm{0}_{3\times3}& H_{O_{r1}} & \bm{0}_{3\times3}
%  \\
%   H_{O_{p1}} &    
%   H_{O_{p2}}
% &  \bm{0}_{3\times3}& \bm{0}_{3\times3}& 
% H_{O_{p3}}
% & H_{O_{p4}}\\
% \end{bmatrix}
% \end{equation}
\begin{small}
\begin{equation} \label{pfH}
    \bm{H}_{pf} = \begin{bmatrix}\bm{H}_{pose}&\bm{H}_{extrinsic}&\bm{0}_{3\times6}\\\bm{H}_{pose1}&\bm{0}_{3\times6}&\bm{H}_{extrinsic1}
    \\\bm{H}_{pose2}&\bm{0}_{3\times6}&\bm{H}_{extrinsic2}
    \end{bmatrix}
\end{equation}
\end{small}
Then we substitute (\ref{pfH}) into (\ref{crlb_formation}) and represent it same as (\ref{purely-LiDARJ}):
\begin{small}
\begin{equation}
    \mathit{\mathbf{J}}_{pf} = \begin{bmatrix}
        \bm{U}+\bm{F} & \bm{B}& \bm{D} \\ \bm{B}^{\top} & \bm{C} &\bm{0}\\ \bm{D}^{\top} &\bm{0}& \bm{E}
    \end{bmatrix}
\end{equation}
\end{small}
Then, we marginalize the odometry-extrinsic-related variable to get the marginalized Fisher Information Matrix of the variables ${\mathit{\mathbf{J}}_{pfmar}}$ as 
\begin{equation}
    {\mathit{\mathbf{J}}_{pfmar}} = \begin{bmatrix}
        \bm{U}+\bm{F}-\bm{D}{\bm{E}^{-1}}\bm{D}^{\top} & \bm{B} \\ \bm{B}^{\top} & \bm{C}
    \end{bmatrix}
\end{equation}
Then following (\ref{crlb1}), the corresponding CRLB for estimated pose is
\begin{equation}
     {\rm CRLB}_{pf} = ({\bm{U} - \bm{B}{\bm{C}^{-1}}\bm{B}^{\top} + \bm{F} - \bm{D}{\bm{E}^{-1}}\bm{D}^{\top}})^{-1}
\end{equation}

3) \textbf{Comparison}:
Considering a case in which only the odometry pose measurement is used (i.e., no extrinsic variables between LiDAR and IMU in the state vector), the Fisher information matrix is
\begin{equation}
    {\mathit{\mathbf{J}}_{op}} = \begin{bmatrix}
        \bm{F} & \bm{D} \\ \bm{D}^{\top} & \bm{E}\\
    \end{bmatrix}
\end{equation}
Under this condition, $\begin{small}({\bm{F} - \bm{D}{\bm{E}^{-1}}\bm{D}^{\top}})^{-1}\end{small}$ is the information matrix corresponding to estimated pose and thus obviously is positive definite. Based on Lemma 1, we can refer that $\!({\bm{F} - \bm{D}{\bm{E}^{-1}}\bm{D}^{\top}}) > 0\!$. According to \cite{fu2021high}, we have
% \begin{align}
%     \begin{split}
%     {\bm{U} - \bm{B}{\bm{C}^{-1}}\bm{B}^{\top} + \bm{F} - \bm{D}{\bm{E}^{-1}}\bm{D}^{\top}} \\
%     > \bm{U} - \bm{B}{\bm{C}^{-1}}\bm{B}^{\top} > 0 
%     \end{split}
% \end{align}
\begin{small}
% \begin{align}
%     {\bm{U} - \bm{B}{\bm{C}^{-1}}\bm{B}^{\top} + \bm{F} - \bm{D}{\bm{E}^{-1}}\bm{D}^{\top}} > \bm{U} - \bm{B}{\bm{C}^{-1}}\bm{B}^{\top} > 0 
% \end{align}
\begin{align}
    \hspace{-2mm}
    \!{({\bm{U}-\bm{B}{\bm{C}^{-1}}\bm{B}^{\top} + \bm{F} - \bm{D}{\bm{E}^{-1}}\bm{D}^{\top}})^{-1}} < (\bm{U} - \bm{B}{\bm{C}^{-1}}\bm{B}^{\top} )^{-1} \!
\end{align}
\end{small}
Then, it can be verified that ${\rm CRLB}_{pf} < {\rm CRLB}_{li}$. Namely, the method that fuses pose measurement and LiDAR points has a smaller lower bound on the covariance of pose estimation than the method that uses LiDAR points alone. Substituting the above analysis into the application scenario, it shows that our sensor-fusion method makes fuller use of the other odometry measurement than those using odometry information as an initial value, and therefore achieves better accuracy in pose estimation.

\begin{figure}
	\setlength{\belowcaptionskip}{-0.6cm}   %调整图片标题与下文距离
	\setlength{\abovecaptionskip}{-0.1cm}
    	\centering \includegraphics[width=0.4\textwidth]{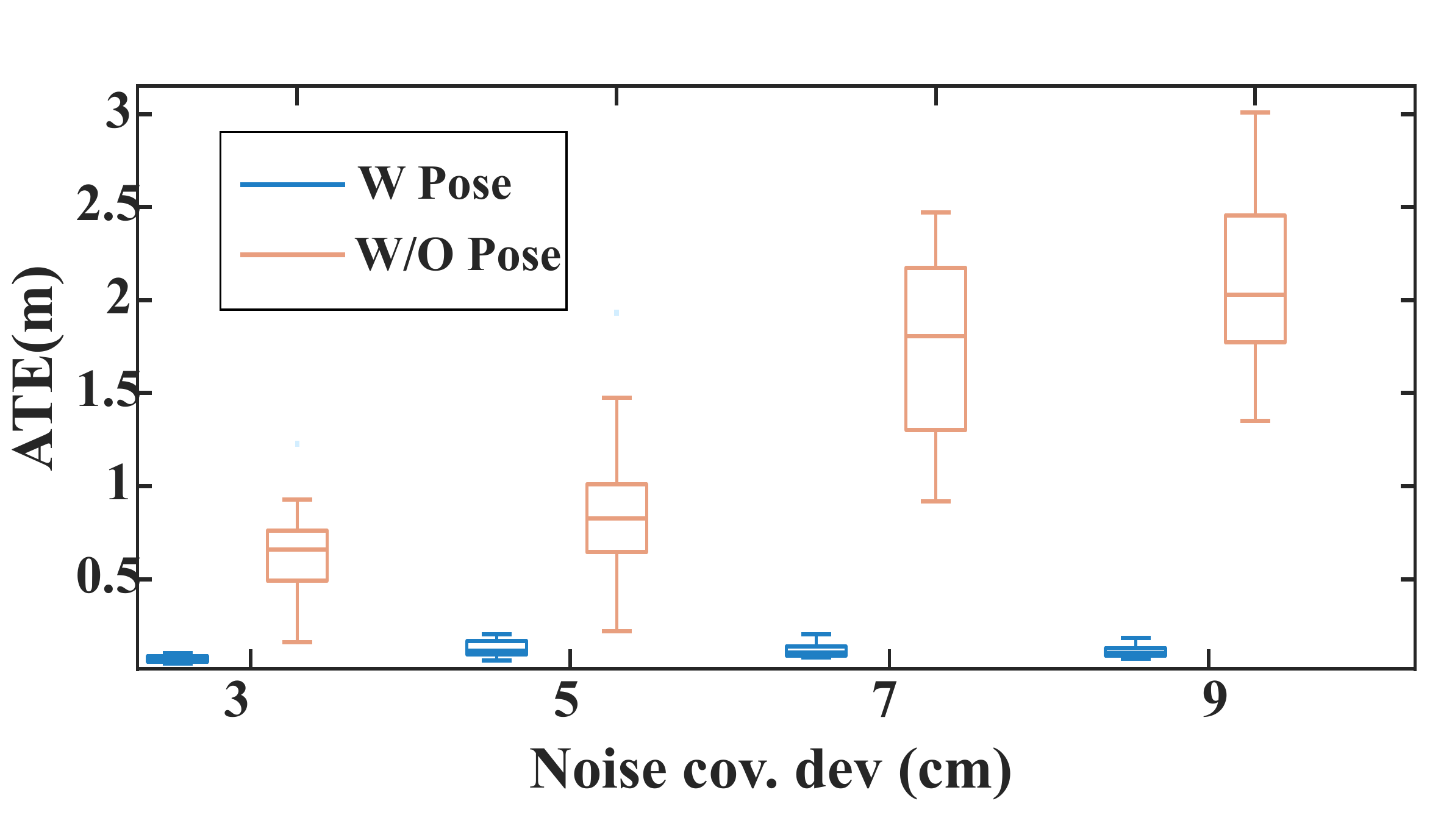}
	\captionsetup{font={footnotesize}}
	\caption{{Simulation validation result. The value of the horizontal coordinate represents the standard deviation of the noise. W and W/O pose mean with and without pose measurement.}}\label{simulation}
\end{figure}
\section{Experiments}
In this section, we fully evaluate the proposed DAMS-LIO method with both simulation and real-world experiments. The simulation experiments are mainly used to verify the advantages of the proposed fusion method. Then we compare the accuracy and robustness of our method against other state-of-the-art methods in the public dataset and demonstrate the computing efficiency.  
\subsection{Simulation Validations}
The primary purpose of the simulation experiment, which is built with the Gazebo simulator, is to validate the theoretical analysis in section \uppercase\expandafter{\romannumeral4}. We construct a long corridor to simulate the degeneration environment and sample data through turtlebot3 mobile robot equipped with the velodyne VLP16 LiDAR sensor and wheel encoder. Similar to the simulator in Openvins \cite{geneva2020openvins}, we obtain the simulated IMU data by sample interpolation of the true values of the robot trajectory.

\begin{figure}[tp] 
	\centering 
% 	\vspace{-0.35cm} %设置与上面正文的距离
	\setlength{\belowcaptionskip}{-0.4cm}   %调整图片标题与下文距离
% 	\subfigtopskip=2pt %设置子图与上面正文或别的内容的距离
	\captionsetup{font={footnotesize}}
% 	\subfigbottomskip=2pt %设置第二行子图与第一行子图的距离，即下面的头与上面的脚的距离
	\subfigcapskip=-5pt %设置子图与子标题之间的距离
	\subfigure[CERBERUS dataset]{
		\label{CERBERUS dataset}
		\includegraphics[width=0.38\textwidth]{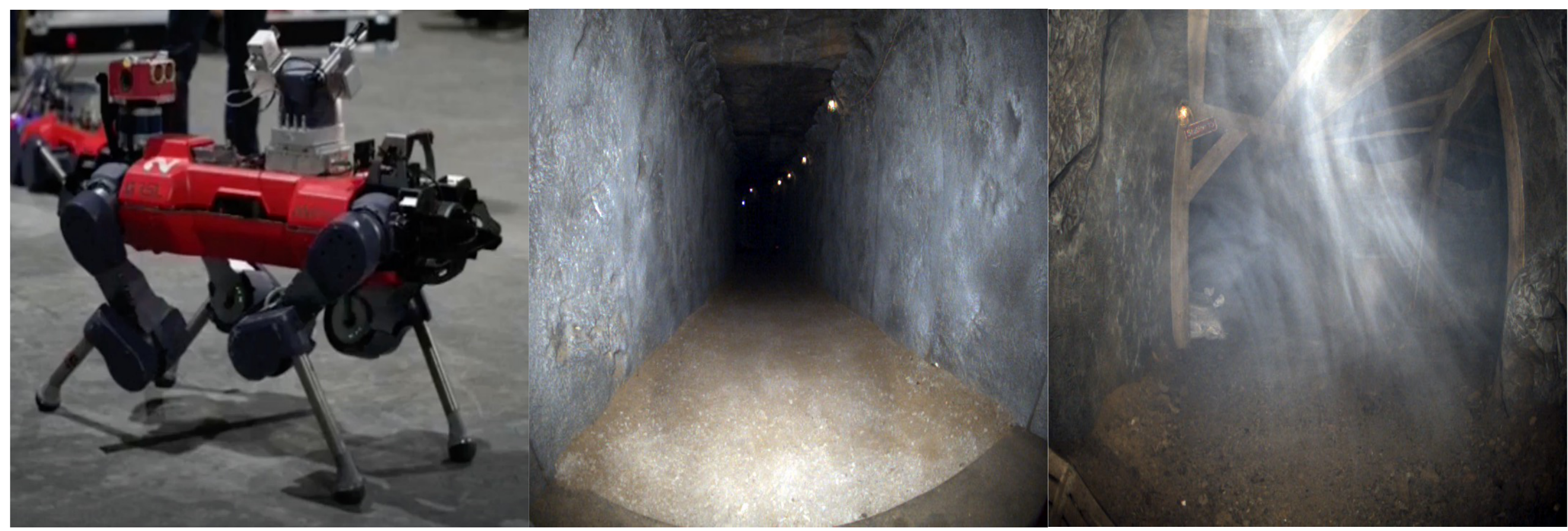}}\vspace{-0.3cm}
    \\
    \subfigure[M2DGR dataset]{
		\label{M2DGR dataset}
		\includegraphics[width=0.38\textwidth]{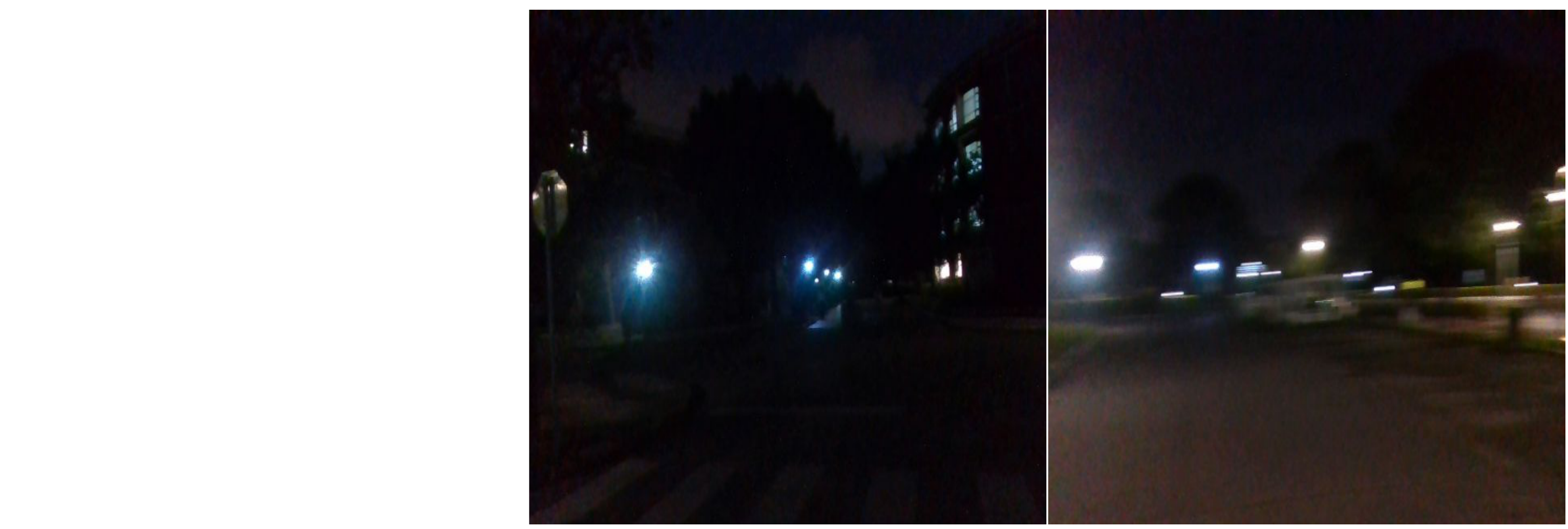}}\vspace{-0.3cm}
	\caption{Illustration of the real-world datasets used in experiment. 
% 	(a) shows complex underground scenes in the CERBERUS dataset, including dusty and long tunnel-like degeneration scenes. (b) shows challenging urban scenes in M2DGR, which contains the poorly illuminated conditions and image-blur caused by sharp turns.
	}
	\label{real-dataset}
\end{figure}
The gaussian noise $\mathit{N}(0,{\sigma}^{2})$ is added to the LiDAR point data. Here the covariance ${\sigma}^{2}$ varies from 3 to 9 cm with an interval of 2 cm. We run our method repeatedly, with and without pose measurement, 20 times to obtain the absolute trajectory error (ATE) and evaluate the mean and dispersion of the errors in the form of box plots. As shown in Fig. \ref{simulation}, with an increase in the noise variance, the mean and covariance of the APE obtained by the method with pose observation are smaller than those without pose measurements. Therefore, this validates that fusing the odometry pose and LiDAR points observation simultaneously can achieve higher accuracy performance and lower covariance bound in pose estimation.
\begin{figure*}[htp]
	\setlength{\belowcaptionskip}{-0.4cm}   %调整图片标题与下文距离
	\centering \includegraphics[width=0.94\textwidth]{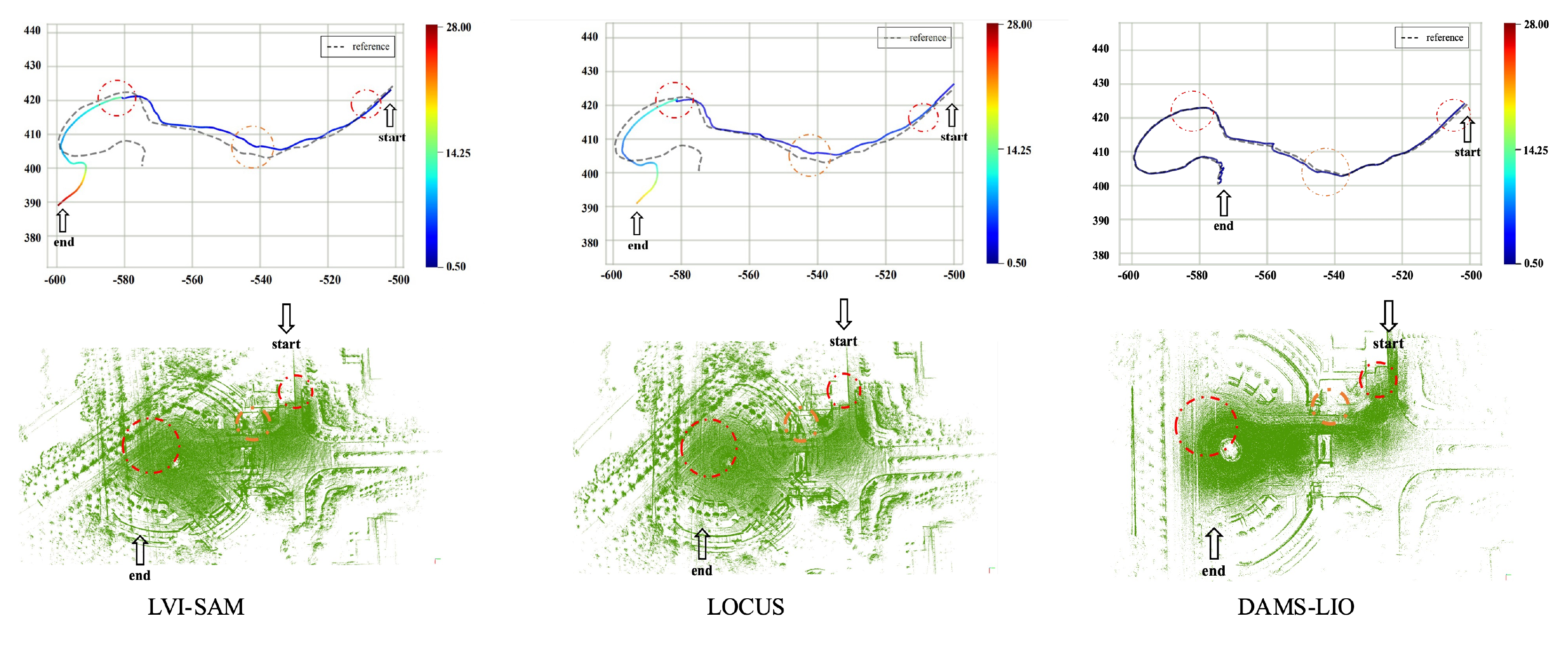} 
	\captionsetup{font={footnotesize}}
	\caption{Map and trajectory comparison of different sensor-fusion methods in the urban data sequence. Red dashed circles indicate limited LiDAR sensing distance, and green dashed means a drop in LiDAR measurement, which severely degrades the performance of LOCUS and LVI-SAM at the end.}
	\label{robustness}
\end{figure*}
% \begin{figure*}[htp] 
% 	\centering 
% 	\vspace{-0.3cm} %设置与上面正文的距离
% 	\captionsetup{font={footnotesize}}
% 	\subfigtopskip=1pt %设置子图与上面正文或别的内容的距离
% % 	\subfigbottomskip=2pt %设置第二行子图与第一行子图的距离，即下面的头与上面的脚的距离
% 	\subfigcapskip=-5pt %设置子图与子标题之间的距离
% 	\subfigure[Trajectory of LVI-SAM]{
% 		\label{Trajectory LVI-SAM}
% 		\includegraphics[width=0.32\linewidth]{fig/map/robust-lvi.pdf}}
% % 	\quad %默认情况下两个子图之间空的较少，使用这个命令加大宽度
% 	\subfigure[Trajectory of LOCUS]{
% 		\label{Trajectory LOCUS}
% 		\includegraphics[width=0.32\linewidth]{fig/map/robust-locus.pdf}}
% 	\subfigure[Trajectory of DAMS-LIO]{
% 		\label{Trajectory DAMS-LIO}
% 		\includegraphics[width=0.32\linewidth]{fig/map/robust-dams.pdf}}
% % 	\quad
%     \\
%     \subfigure[mapping result of LVI-SAM]{
% 		\label{MAP LVI-SAM}
% 		\includegraphics[width=0.32\linewidth]{fig/map/map-lvisam.png}}
% % 	\quad %默认情况下两个子图之间空的较少，使用这个命令加大宽度
% 	\subfigure[mapping result of LOCUS]{
% 		\label{MAP LOCUS}
% 		\includegraphics[width=0.32\linewidth]{fig/map/map-locus.png}}
% 	\subfigure[mapping result of DAMS-LIO]{
% 		\label{MAP DAMS-LIO}
% 		\includegraphics[width=0.32\linewidth]{fig/map/map-dams.png}}
% % 	\quad
% 	\caption{Map and trajectory comparison of LVI-SAM, LOCUS and DAMS-LIO in the urban data sequence}
% 	\label{robustness}
% \end{figure*}
\subsection{Real-world Experiments}
To further validate the practical performance of our method, we compare it with current state-of-the-art state estimation systems on publicly available and challenging datasets.
% To maintain the viability of various algorithm operations, we choose the dataset collected on-board the four ANYmal C robots used by Team CERBERUS in their winning run in the DARPA Subterranean Challenge finals \cite{tranzatto2022cerberus} and a multi-sensor and multi-scenario SLAM dataset, M2DGR\cite{yin2021m2dgr}.
% As shown in Fig \ref{real-dataset}, these datasets contain a variety of challenging scenarios, including structure-less and poor-illumination environments(e.g. long tunnels, dusty scene, image-blur). Also these datasets have a rich set of sensor types, including camera image, IMU measurements, LiDAR measurements, so that we can evaluate various types of algorithms .
To maintain the viability of various methods operations, we choose the CERBERUS DARPA Subterranean Challenge dataset \cite{tranzatto2022cerberus} and the M2DGR dataset Since these scenarios do not have degeneration of LIO, we select some periods in each scenario to limit the range of LiDAR measurements to produce degeneration. In addition, for gate03, we also added a period of time when the LiDAR data was lost. \cite{yin2021m2dgr}, which have a rich set of sensor types as shown in Fig \ref{real-dataset}.

% \begin{figure}[htp]
% 	\setlength{\belowcaptionskip}{-0.5cm}   %调整图片标题与下文距离
% 	\setlength{\abovecaptionskip}{-0.2cm}
% 	\centering \includegraphics[width=0.45\textwidth]{fig/frame.pdf}
% 	\captionsetup{font={footnotesize}}
% 	\caption{{}}\label{real-dataset}
% \end{figure}

1) \textbf{Accuracy}:
We first use these challenging datasets to evaluate the accuracy of state estimation. We compare our method with other vision-based methods (VINS-MONO \cite{qin2018vins}), LiDAR-based method (LIO-SAM \cite{shan2020lio}, Fast-LIO2), and sensor-fusion methods (LVI-SAM \cite{shan2021lvi}, LOCUS \cite{palieri2020locus}). We use the trajectory output by VINS-MONO as the odometer input for LOCUS and our method. The EVO package \cite{grupp2017evo} is adopted to calculate the translational part of ATE against the ground truth. The maximum and mean results of ATEs are presented in Table \ref{ate compare}.
It can be seen that our method has lower trajectory error and lower error fluctuation range in most scenarios. Besides, when there is a large error in the odometry (i.e., the max error of VINS-MONO is large), methods such as LOCUS that use the odometry as the initial matching value are more likely to fail. In contrast, the proposed method still maintains good accuracy. 
For some scenarios, the degraded environment does not cause a large error in the LiDAR odometer (e.g. long corridors can be swept to the end), when the introduction of other odometers with larger errors will reduce the accuracy of our algorithm, such as anymal1.

% The table shows that our method has lower trajectory error and lower error fluctuation range in each scenario. 
% Also, it can be seen that LIO-SAM, which does not utilize odometer information, is more likely to fail in the CERBERUS dataset where a large number of degeneration scenarios exist. 
% In addition, when there is a large error in the odometer (i.e. the max error of VINS-MONO is large), methods such as LOCUS that use the odometer as the matching initial value are more likely to fail, while the proposed method still maintains a good accuracy.

2) \textbf{Robustness}:
Our critical insight is that our method can have accuracy estimation relying on the fusion of LiDAR and pose measurement even with poor observations of other odometry and LiDAR points, which highlights the robustness of systems. Therefore, we select gate03 in dataset M2DGR to manually add some difficult scenarios, including the restricted maximum LiDAR sensing distance of 5m in the 15-20s and 165-180s time periods and the loss of LiDAR data at 100-105s (the part in the red and green dashed circle in Fig. \ref{robustness}).
The comparison of trajectory accuracy and map-building results of the three different methods is shown in Fig \ref{robustness}.
It can be seen that compared to LVI-SAM and LOCUS, our method is less sensitive to sensor data drop and poor sensor data. In addition, compared to LOCUS, since we consider both odometry pose and LiDAR point observations in the update process, it allows us to guarantee the accuracy of the estimation despite the poor data from LiDAR (limited measurement range) and poor odometry pose input (The result of VINS-MONO is poor at night). While methods like LOCUS, which only use the odometry as the initial value, will produce large errors and affect the map-building results.
% Then we use the poor vision raw data or estimation result as input for our method and other two sensor-fusion method. Besides, we also limit the measurement range of LiDAR in certain time periods to increase the difficulty of test. 
\begin{figure}
	\setlength{\belowcaptionskip}{-0.9cm}   %调整图片标题与下文距离
	\centering \includegraphics[width=0.32\textwidth]{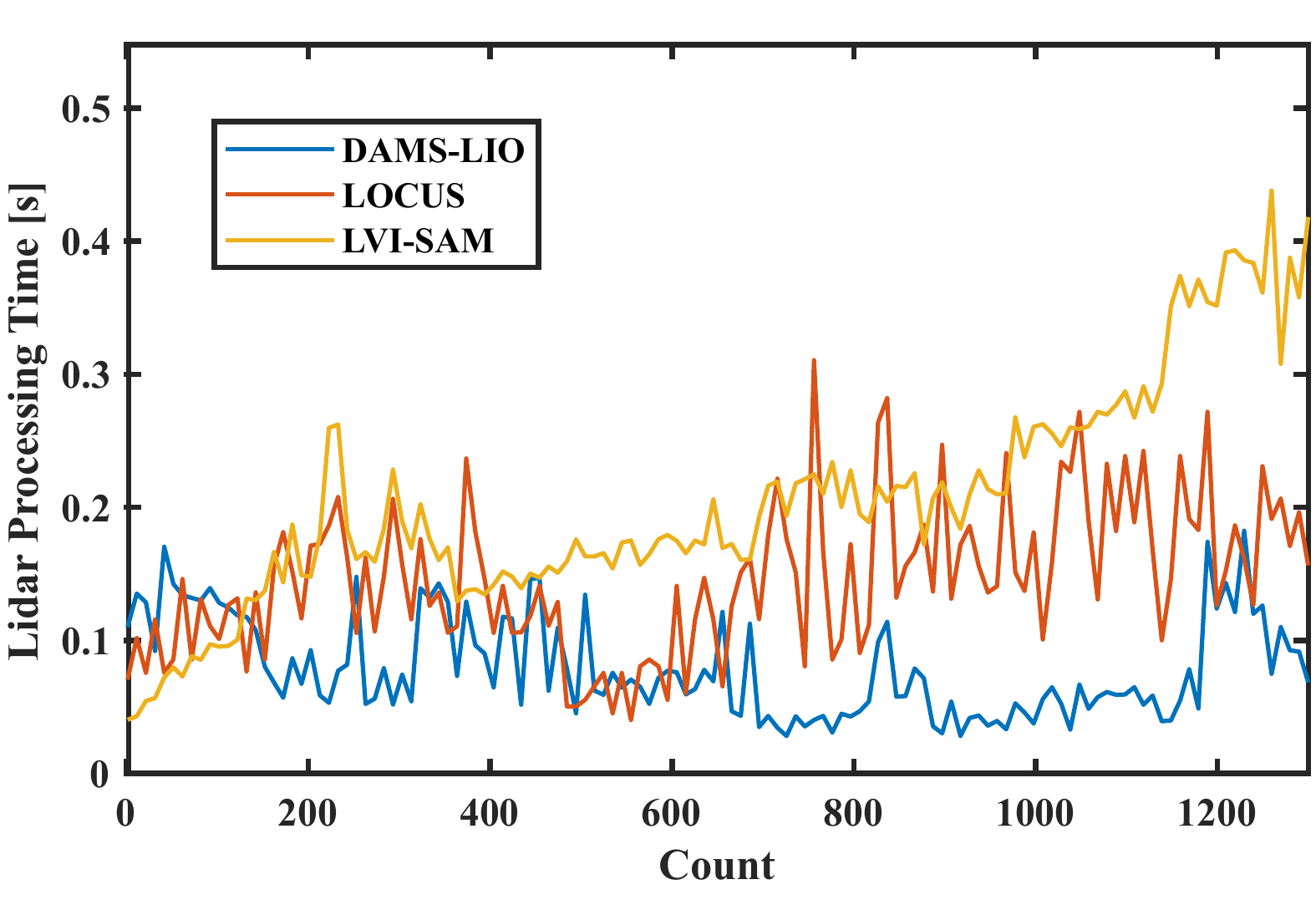}
	\captionsetup{font={footnotesize}}
	\caption{{Comparison of the running time of each system.}}\label{running time}
\end{figure}

3) \textbf{Efficiency}:
We compare the computation time taken by different methods to process the LiDAR data in operation, starting from the reception of LiDAR data until the pose is estimated. All efficiency tests are performed on a laptop ( intel i5-9300 @2.4GHz$\times$8 ) running ubuntu 16.04 LTS. As shown in Fig \ref{running time}, our method can always maintain a lower runtime than other methods, with the runtime growth.
Since the LiDAR data is published at 10 Hz, and the average processing time of our method in Fig \ref{running time} is mostly less than 0.1s, which indicates that our system achieves an efficiency of real-time operation.
\section{Conclusions}
Multi-sensor fusion is a promising solution to the problem of state estimation in the perceptually-challenge environment while adopting a fusion scheme that can balance robustness and accuracy is intractable.
To handle this problem, we propose the DAMS-LIO, a degeneration-aware and modular sensor-fusion LiDAR-inertial odometry system that incorporates both LiDAR and odometry pose measurements when degeneration is detected in LIO system. The CRLB theory analysis and extensive experiments are performed to demonstrate that our method has high accuracy, robustness, and efficiency. 
% The CRLB theory is used to confirm the advantage of this fusion scheme. To facilitate the deployment of the algorithm, we also estimate the extrinsic parameters between the other odometry and LIO.
% The proposed method is evaluated in both simulation and real-world experiments. The results show that our is able to achieve more accurate pose estimation results in extreme environment and shows high robustness especially when the pose measurement provided by other odometry is poor.
% In addition, the efficiency of the algorithm is evaluated to demonstrate the possibility of deploying our algorithm to robotic platforms with limited computing resources.
In future work, we intend to evaluate the observability information obtained from the odometry and retain only the most informative segments to improve accuracy and robustness.

% \addtolength{\textheight}{-12cm}   % This command serves to balance the column lengths
                                  % on the last page of the document manually. It shortens
                                  % the textheight of the last page by a suitable amount.
                                  % This command does not take effect until the next page
                                  % so it should come on the page before the last. Make
                                  % sure that you do not shorten the textheight too much.

%%%%%%%%%%%%%%%%%%%%%%%%%%%%%%%%%%%%%%%%%%%%%%%%%%%%%%%%%%%%%%%%%%%%%%%%%%%%%%%%

%%%%%%%%%%%%%%%%%%%%%%%%%%%%%%%%%%%%%%%%%%%%%%%%%%%%%%%%%%%%%%%%%%%%%%%%%%%%%%%%

%\begin{thebibliography}{99}
%
%\bibitem{c1} G. O
%
%
%
%
%
%
%\end{thebibliography}
%
\bibliographystyle{IEEEtran}
\bibliography{ref}

% Generated by IEEEtran.bst, version: 1.14 (2015/08/26)
\begin{thebibliography}{10}
\providecommand{\url}[1]{#1}
\csname url@samestyle\endcsname
\providecommand{\newblock}{\relax}
\providecommand{\bibinfo}[2]{#2}
\providecommand{\BIBentrySTDinterwordspacing}{\spaceskip=0pt\relax}
\providecommand{\BIBentryALTinterwordstretchfactor}{4}
\providecommand{\BIBentryALTinterwordspacing}{\spaceskip=\fontdimen2\font plus
\BIBentryALTinterwordstretchfactor\fontdimen3\font minus
  \fontdimen4\font\relax}
\providecommand{\BIBforeignlanguage}[2]{{%
\expandafter\ifx\csname l@#1\endcsname\relax
\typeout{** WARNING: IEEEtran.bst: No hyphenation pattern has been}%
\typeout{** loaded for the language `#1'. Using the pattern for}%
\typeout{** the default language instead.}%
\else
\language=\csname l@#1\endcsname
\fi
#2}}
\providecommand{\BIBdecl}{\relax}
\BIBdecl

\bibitem{erdelj2017help}
M.~Erdelj, E.~Natalizio, K.~R. Chowdhury, and I.~F. Akyildiz, ``Help from the
  sky: Leveraging uavs for disaster management,'' \emph{IEEE Pervasive
  Computing}, vol.~16, no.~1, pp. 24--32, 2017.

\bibitem{pan2020gem}
Y.~Pan, X.~Xu, X.~Ding, S.~Huang, Y.~Wang, and R.~Xiong, ``Gem: online globally
  consistent dense elevation mapping for unstructured terrain,'' \emph{IEEE
  Transactions on Instrumentation and Measurement}, vol.~70, pp. 1--13, 2020.

\bibitem{jiao20192}
Y.~Jiao, Y.~Wang, B.~Fu, X.~Ding, Q.~Tan, L.~Chen, and R.~Xiong, ``2-entity
  ransac for robust visual localization in changing environment,'' in
  \emph{2019 IEEE/RSJ International Conference on Intelligent Robots and
  Systems (IROS)}.\hskip 1em plus 0.5em minus 0.4em\relax IEEE, 2019, pp.
  2478--2485.

\bibitem{jiao2021robust}
Y.~Jiao, L.~Liu, B.~Fu, X.~Ding, M.~Wang, Y.~Wang, and R.~Xiong, ``Robust
  localization for planar moving robot in changing environment: A perspective
  on density of correspondence and depth,'' in \emph{2021 IEEE International
  Conference on Robotics and Automation (ICRA)}.\hskip 1em plus 0.5em minus
  0.4em\relax IEEE, 2021, pp. 4006--4012.

\bibitem{shan2018lego}
T.~Shan and B.~Englot, ``Lego-loam: Lightweight and ground-optimized lidar
  odometry and mapping on variable terrain,'' in \emph{2018 IEEE/RSJ
  International Conference on Intelligent Robots and Systems (IROS)}.\hskip 1em
  plus 0.5em minus 0.4em\relax IEEE, 2018, pp. 4758--4765.

\bibitem{zhao2021super}
S.~Zhao, H.~Zhang, P.~Wang, L.~Nogueira, and S.~Scherer, ``Super odometry:
  Imu-centric lidar-visual-inertial estimator for challenging environments,''
  in \emph{2021 IEEE/RSJ International Conference on Intelligent Robots and
  Systems (IROS)}.\hskip 1em plus 0.5em minus 0.4em\relax IEEE, 2021, pp.
  8729--8736.

\bibitem{jing2022dxq}
X.~Jing, X.~Ding, R.~Xiong, H.~Deng, and Y.~Wang, ``Dxq-net: differentiable
  lidar-camera extrinsic calibration using quality-aware flow,'' in \emph{2022
  IEEE/RSJ International Conference on Intelligent Robots and Systems
  (IROS)}.\hskip 1em plus 0.5em minus 0.4em\relax IEEE, 2022, pp. 6235--6241.

\bibitem{shan2021lvi}
T.~Shan, B.~Englot, C.~Ratti, and D.~Rus, ``Lvi-sam: Tightly-coupled
  lidar-visual-inertial odometry via smoothing and mapping,'' in \emph{2021
  IEEE international conference on robotics and automation (ICRA)}.\hskip 1em
  plus 0.5em minus 0.4em\relax IEEE, 2021, pp. 5692--5698.

\bibitem{zuo2019lic}
X.~Zuo, P.~Geneva, W.~Lee, Y.~Liu, and G.~Huang, ``Lic-fusion:
  Lidar-inertial-camera odometry,'' in \emph{2019 IEEE/RSJ International
  Conference on Intelligent Robots and Systems (IROS)}.\hskip 1em plus 0.5em
  minus 0.4em\relax IEEE, 2019, pp. 5848--5854.

\bibitem{lin2021r}
J.~Lin, C.~Zheng, W.~Xu, and F.~Zhang, ``R $^{2}$ live: A robust, real-time,
  lidar-inertial-visual tightly-coupled state estimator and mapping,''
  \emph{IEEE Robotics and Automation Letters}, vol.~6, no.~4, pp. 7469--7476,
  2021.

\bibitem{palieri2020locus}
M.~Palieri, B.~Morrell, A.~Thakur, K.~Ebadi, J.~Nash, A.~Chatterjee,
  C.~Kanellakis, L.~Carlone, C.~Guaragnella, and A.-a. Agha-Mohammadi, ``Locus:
  A multi-sensor lidar-centric solution for high-precision odometry and 3d
  mapping in real-time,'' \emph{IEEE Robotics and Automation Letters}, vol.~6,
  no.~2, pp. 421--428, 2020.

\bibitem{khattak2020complementary}
S.~Khattak, H.~Nguyen, F.~Mascarich, T.~Dang, and K.~Alexis, ``Complementary
  multi--modal sensor fusion for resilient robot pose estimation in
  subterranean environments,'' in \emph{2020 International Conference on
  Unmanned Aircraft Systems (ICUAS)}.\hskip 1em plus 0.5em minus 0.4em\relax
  IEEE, 2020, pp. 1024--1029.

\bibitem{rouvcek2019darpa}
T.~Rou{\v{c}}ek, M.~Pecka, P.~{\v{C}}{\'\i}{\v{z}}ek,
  T.~Pet{\v{r}}{\'\i}{\v{c}}ek, J.~Bayer, V.~{\v{S}}alansk{\`y}, D.~He{\v{r}}t,
  M.~Petrl{\'\i}k, T.~B{\'a}{\v{c}}a, V.~Spurn{\`y} \emph{et~al.}, ``Darpa
  subterranean challenge: Multi-robotic exploration of underground
  environments,'' in \emph{International Conference on Modelling and Simulation
  for Autonomous Systems}.\hskip 1em plus 0.5em minus 0.4em\relax Springer,
  2019, pp. 274--290.

\bibitem{xu2022fast}
W.~Xu, Y.~Cai, D.~He, J.~Lin, and F.~Zhang, ``Fast-lio2: Fast direct
  lidar-inertial odometry,'' \emph{IEEE Transactions on Robotics}, 2022.

\bibitem{gorman1990lower}
J.~D. Gorman and A.~O. Hero, ``Lower bounds for parametric estimation with
  constraints,'' \emph{IEEE Transactions on Information Theory}, vol.~36,
  no.~6, pp. 1285--1301, 1990.

\bibitem{zhang2018laser}
J.~Zhang and S.~Singh, ``Laser--visual--inertial odometry and mapping with high
  robustness and low drift,'' \emph{Journal of field robotics}, vol.~35, no.~8,
  pp. 1242--1264, 2018.

\bibitem{yang2019degenerate}
Y.~Yang, P.~Geneva, K.~Eckenhoff, and G.~Huang, ``Degenerate motion analysis
  for aided ins with online spatial and temporal sensor calibration,''
  \emph{IEEE Robotics and Automation Letters}, vol.~4, no.~2, pp. 2070--2077,
  2019.

\bibitem{li2013high}
M.~Li and A.~I. Mourikis, ``High-precision, consistent ekf-based
  visual-inertial odometry,'' \emph{The International Journal of Robotics
  Research}, vol.~32, no.~6, pp. 690--711, 2013.

\bibitem{mourikis2007multi}
A.~I. Mourikis, S.~I. Roumeliotis \emph{et~al.}, ``A multi-state constraint
  kalman filter for vision-aided inertial navigation.'' in \emph{ICRA},
  vol.~2.\hskip 1em plus 0.5em minus 0.4em\relax Citeseer, 2007, p.~6.

\bibitem{sola2018micro}
J.~Sola, J.~Deray, and D.~Atchuthan, ``A micro lie theory for state estimation
  in robotics,'' \emph{arXiv preprint arXiv:1812.01537}, 2018.

\bibitem{he2021kalman}
D.~He, W.~Xu, and F.~Zhang, ``Kalman filters on differentiable manifolds,''
  \emph{arXiv preprint arXiv:2102.03804}, 2021.

\bibitem{ding2021degeneration}
X.~Ding, F.~Han, T.~Yang, Y.~Wang, and R.~Xiong, ``Degeneration-aware
  localization with arbitrary global-local sensor fusion,'' \emph{Sensors},
  vol.~21, no.~12, p. 4042, 2021.

\bibitem{domhof2017multi}
J.~Domhof, R.~Happee, and P.~Jonker, ``Multi-sensor object tracking performance
  limits by the cramer-rao lower bound,'' in \emph{2017 20th International
  Conference on Information Fusion (Fusion)}.\hskip 1em plus 0.5em minus
  0.4em\relax IEEE, 2017, pp. 1--8.

\bibitem{blanc2007data}
C.~Blanc, P.~Checchin, S.~Gidel, and L.~Trassoudaine, ``Data fusion performance
  evaluation for range measurements combined with cartesian ones for road
  obstacle tracking,'' in \emph{2007 IEEE International Conference on Vehicular
  Electronics and Safety}.\hskip 1em plus 0.5em minus 0.4em\relax IEEE, 2007,
  pp. 1--6.

\bibitem{kowalski2019crlb}
M.~Kowalski, Y.~Bar-Shalom, P.~Willett, B.~Milgrom, and R.~Ben-Dov, ``Crlb for
  multi-sensor rotational bias estimation for passive sensors without target
  state estimation,'' in \emph{Signal Processing, Sensor/Information Fusion,
  and Target Recognition XXVIII}, vol. 11018.\hskip 1em plus 0.5em minus
  0.4em\relax SPIE, 2019, pp. 33--41.

\bibitem{horn2012matrix}
R.~A. Horn and C.~R. Johnson, \emph{Matrix analysis}.\hskip 1em plus 0.5em
  minus 0.4em\relax Cambridge university press, 2012.

\bibitem{fu2021high}
B.~Fu, F.~Han, Y.~Wang, Y.~Jiao, X.~Ding, Q.~Tan, L.~Chen, M.~Wang, and
  R.~Xiong, ``High-precision multicamera-assisted camera-imu calibration:
  Theory and method,'' \emph{IEEE Transactions on Instrumentation and
  Measurement}, vol.~70, pp. 1--17, 2021.

\bibitem{geneva2020openvins}
P.~Geneva, K.~Eckenhoff, W.~Lee, Y.~Yang, and G.~Huang, ``Openvins: A research
  platform for visual-inertial estimation,'' in \emph{2020 IEEE International
  Conference on Robotics and Automation (ICRA)}.\hskip 1em plus 0.5em minus
  0.4em\relax IEEE, 2020, pp. 4666--4672.

\bibitem{tranzatto2022cerberus}
M.~Tranzatto, T.~Miki, M.~Dharmadhikari, L.~Bernreiter, M.~Kulkarni,
  F.~Mascarich, O.~Andersson, S.~Khattak, M.~Hutter, R.~Siegwart \emph{et~al.},
  ``Cerberus in the darpa subterranean challenge,'' \emph{Science Robotics},
  vol.~7, no.~66, p. eabp9742, 2022.

\bibitem{yin2021m2dgr}
J.~Yin, A.~Li, T.~Li, W.~Yu, and D.~Zou, ``M2dgr: A multi-sensor and
  multi-scenario slam dataset for ground robots,'' \emph{IEEE Robotics and
  Automation Letters}, vol.~7, no.~2, pp. 2266--2273, 2021.

\bibitem{qin2018vins}
T.~Qin, P.~Li, and S.~Shen, ``Vins-mono: A robust and versatile monocular
  visual-inertial state estimator,'' \emph{IEEE Transactions on Robotics},
  vol.~34, no.~4, pp. 1004--1020, 2018.

\bibitem{shan2020lio}
T.~Shan, B.~Englot, D.~Meyers, W.~Wang, C.~Ratti, and D.~Rus, ``Lio-sam:
  Tightly-coupled lidar inertial odometry via smoothing and mapping,'' in
  \emph{2020 IEEE/RSJ international conference on intelligent robots and
  systems (IROS)}.\hskip 1em plus 0.5em minus 0.4em\relax IEEE, 2020, pp.
  5135--5142.

\bibitem{grupp2017evo}
M.~Grupp, ``evo: Python package for the evaluation of odometry and slam.''
  \url{https://github.com/MichaelGrupp/evo}, 2017.

\end{thebibliography}

\end{document}